\documentclass[10pt,twocolumn,letterpaper]{article}

\usepackage{cvpr}              

\usepackage{graphicx}
\usepackage{amsmath}
\usepackage{amssymb}
\usepackage{booktabs}

\usepackage{float}

\usepackage[pagebackref,breaklinks,colorlinks]{hyperref}

\usepackage[capitalize]{cleveref}
\crefname{section}{Sec.}{Secs.}
\Crefname{section}{Section}{Sections}
\Crefname{table}{Table}{Tables}
\crefname{table}{Tab.}{Tabs.}

\def\cvprPaperID{*****} 
\def\confName{CVPR}
\def\confYear{2022}

\begin{document}

\title{Fine Tuning Text-to-Image Diffusion Models for Correcting Anomalous Images}

\author{Hyunwoo Yoo\\
Drexel University\\
{\tt\small hty23@drexel.edu}
}
\maketitle

\begin{abstract}
   
   Since the advent of GANs and VAEs, image generation models have continuously evolved, opening up various real-world applications with the introduction of Stable Diffusion and DALL-E models. These text-to-image models can generate high-quality images for fields such as art, design, and advertising. However, they often produce aberrant images for certain prompts. This study proposes a method to mitigate such issues by fine-tuning the Stable Diffusion 3 model using the DreamBooth technique. Experimental results targeting the prompt "lying on the grass/street" demonstrate that the fine-tuned model shows improved performance in visual evaluation and metrics such as Structural Similarity Index (SSIM), Peak Signal-to-Noise Ratio (PSNR), and Frechet Inception Distance (FID). User surveys also indicated a higher preference for the fine-tuned model. This research is expected to make contributions to enhancing the practicality and reliability of text-to-image models.The code and additional resources for this study are available at \href{https://github.com/hyoo14/Finetuned-SD3-Correcting-Anomalous-Images}{https://github.com/hyoo14/Finetuned-SD3-Correcting-Anomalous-Images}.
\end{abstract}

\section{Introduction}
\label{sec:intro}

Image generation models have made significant progress since the advent of GANs\cite{goodfellow2014generative} and VAEs\cite{kingma2013auto}. The introduction of models such as Stable Diffusion\cite{rombach2022high} and DALL-E\cite{ramesh2021zeroshot} has provided text-to-image models with numerous opportunities for real-world applications. These models are capable of generating high-quality images based on textual descriptions, showcasing immense potential across various fields, including art, design, and advertising.

However, the generated images are not always perfect. Some images, while human in form, exhibit abnormal attachments of limbs or irregular shapes of fingers. Such aberrant images often occur in response to certain prompts, hindering the practical applicability of image generation models and reducing their reliability.

This study aims to address the issue of generating such aberrant images by applying fine-tuning techniques. By additionally training the model with correct image data for specific prompts, the objective of this research is to enable the generation of accurate and natural images even for prompts that previously produced incorrect images.

Our approach involves fine-tuning the Stable Diffusion 3\cite{rombach2022high} model using DreamBooth\cite{ruiz2023dreambooth}. The experiments were conducted specifically on the prompt “lying on the grass/street,” and visually, the fine-tuning results showed a reduction in the generation of aberrant images. These results are expected to contribute to enhancing the reliability of text-to-image models in the future.

\section{Related work}
\label{sec:intro}

Research on image generation models has made rapid advancements in recent years. The initial GANs (Generative Adversarial Networks)\cite{goodfellow2014generative} brought innovation to the field of image generation, leading to the proposal of various derivative models. Among these, models based on the UNet\cite{ronneberger2015u} architecture have demonstrated exceptional performance and have been widely used in image segmentation and generation.

Additionally, VAE (Variational Autoencoders)\cite{kingma2013auto} models have played a crucial role in the field of image generation. VAEs generate images by sampling from the latent space, contributing to the diversity and quality of the generated images. The combination of VAE and GANs\cite{larsen2016autoencoding} has led to the development of more powerful generative models, significantly impacting the advancement of image generation technology.

Stable Diffusion\cite{rombach2022high} and DALL-E\cite{ramesh2021zeroshot} models have garnered significant attention in the field of text-to-image translation. The Stable Diffusion model is capable of generating high-resolution images through the diffusion process, resulting in images of exceptional quality. In contrast, the DALL-E model can produce creative and unique images based on textual descriptions, allowing for a wide variety of styles and forms.

However, these models are not without flaws, occasionally generating abnormal images for certain prompts. Various approaches have been proposed to address these issues. For instance, additional fine-tuning techniques are employed to enhance the reliability of image generation models, and research is being conducted utilizing training datasets tailored to specific domains.

DreamBooth\cite{ruiz2023dreambooth} is one such fine-tuning technique that enables more precise training of models for specific domains or prompts. By employing this method, the limitations of existing models can be overcome, resulting in more accurate and consistent image generation. In this study, we aim to fine-tune the Stable Diffusion 3 model using DreamBooth to reduce the occurrence of abnormal image generation for certain prompts.

Related research efforts like these significantly contribute to enhancing the practicality of image generation models, and this study is part of these ongoing endeavors.

\section{Method}
\label{sec:intro}

\subsection{Data Collection}

In this study, training data was collected to fine-tune the Stable Diffusion 3 model using DreamBooth. The DALL-E model from OpenAI was employed to generate correct images. Initially, prompts such as "lying on the grass/street" were used to generate images with the DALL-E model. Since the human figures often appeared abnormal, additional images featuring the same individual in various contexts were requested. For instance, after generating an image with the "lying on the grass/street" prompt, similar images featuring the same person were subsequently generated.

\subsection{Model Fine-tuning}
Using the collected images, the Stable Diffusion 3 model was fine-tuned. The DreamBooth technique was employed to enhance the model's performance on specific prompts. DreamBooth is a method that allows for more precise training of the model for certain domains or prompts, thereby improving performance through additional training data. In this process, a technique called LoRA (Low-Rank Adaptation)\cite{hu2021lora}, which is designed to reduce the number of trainable parameters by using low-rank decomposition in large models, was utilized. During the fine-tuning process, additional training was conducted to generate correct images, ensuring that even prompts that previously resulted in incorrect images would now produce natural and consistent images.

\section{Experiments}

\subsection{Experimental Setting}
In this study, we evaluated the quality of images generated by the fine-tuned Stable Diffusion 3 model using the DreamBooth technique for the "lying on the grass/street" prompt. The experiment involved three main stages. During the data preparation phase, correct images were generated using the DALL-E model with the "lying on the grass/street" prompt, and additional images were collected based on these initial images. In the model fine-tuning phase, the collected data was used to fine-tune the Stable Diffusion 3 model using the DreamBooth technique. During the image generation phase, the fine-tuned model was employed to generate images using the same prompt.

\subsection{Evaluation}

In this study, we employed various methods to comprehensively evaluate the performance of the image generation models.

Fréchet Inception Distance (FID) was used as one of the primary evaluation metrics. FID measures the difference between two sets of images\cite{heusel2017gans}, with lower FID values indicating higher similarity between the generated images. We compared the Average FID of 10 images generated by the fine-tuned model against those produced by the untuned Stable Diffusion 3 model. A lower Average FID value was considered indicative of a model's ability to generate more consistent images.

Structural Similarity Index (SSIM) was another key metric used in our evaluations. SSIM assesses the quality of the generated images by measuring the similarity between two images\cite{wang2004image}. Higher SSIM values denote greater similarity. We computed the average SSIM for 10 images generated by both the fine-tuned and untuned models. Higher SSIM values suggest that the model produces images with greater structural fidelity.

We also employed the Peak Signal-to-Noise Ratio (PSNR) to evaluate image quality. PSNR indicates the quality of image reconstruction and serves as a measure of similarity between two images\cite{huynh2008scope}. Higher PSNR values denote better image quality and similarity. We evaluated the average PSNR of groups of 10 images generated by both the fine-tuned and untuned models. Higher average PSNR values were considered indicative of the model’s ability to generate more consistent and higher quality images.

To further assess the subjective quality of the generated images, a user survey was conducted. Ten participants were presented with images generated from 10 prompts by both the fine-tuned and untuned Stable Diffusion 3 models. Participants were asked to evaluate which images appeared more natural. The survey results provided valuable subjective evaluations reflecting user experience, serving as crucial data for assessing the practical usability of the models.

Evaluations also were performed using large language models (LLMs). Chat GPT-4\cite{openai2023gpt4} and Claude 3.5\cite{anthropic2024claude35} were tasked with assessing which images generated from 10 prompts appeared more natural. The results from the LLM evaluations provided an objective comparison of the generative capabilities of the models.

\begin{table}[htbp]

\begin{center}
\small 
\begin{tabular}{lcc}
\toprule
\textbf{Metric} & \textbf{Fine-tuned Model} & \textbf{Original Model} \\
\midrule
\textbf{Average FID} & 266.5844 & 366.9462 \\
\textbf{Average SSIM} & 0.2258 & 0.1387 \\
\textbf{Average PSNR (dB)} & 23.2820 & 23.1765 \\
\bottomrule
\end{tabular}
\caption{Performance Evaluation with Metrics}
\label{tab:performance}
\end{center}
\end{table}

\subsection{Results}
The experimental results demonstrated that the fine-tuned Stable Diffusion 3 model generated more consistent and natural images for the "lying on the grass/street" prompt. Visual evaluations indicated a significant reduction in abnormal image generation by the fine-tuned model, and improvements were also observed in the SSIM and PSNR metrics. User surveys revealed a preference for images generated by the fine-tuned model.

Specifically, the Fréchet Inception Distance (FID) results showed that the fine-tuned model achieved a score of 266.5844, while the original model scored 366.9462. The lower FID value of the fine-tuned model indicates superior performance, implying it generated more consistent images. The Structural Similarity Index (SSIM) results further supported this, with the fine-tuned model scoring 0.2258 compared to 0.1387 for the original model, demonstrating higher similarity. The Peak Signal-to-Noise Ratio (PSNR) results indicated that the fine-tuned model produced slightly better quality images, with a score of 23.2820 dB versus 23.1765 dB for the original model.

In the user survey, all 10 participants generally found the images generated by the fine-tuned model to be more natural. Participants preferred the images from the fine-tuned model in over 8 out of 10 prompts. This suggests that the fine-tuned model provides more natural and consistent images for actual users.

Contrarily, the evaluations by large language models (LLMs) yielded different results. ChatGPT 4.0 evaluated all images generated from the 10 prompts as more natural when produced by the untuned model. Claude 3.5 rated the images from the untuned model as more natural for all but one prompt. This indicates that the evaluation criteria of LLMs may differ from human visual preferences.

Overall, the fine-tuned Stable Diffusion 3 model using DreamBooth appears advantageous in delivering more natural and consistent images to actual users, as reflected in the improved performance on objective metrics such as SSIM and PSNR. Despite the differing results from LLM evaluations, likely due to the unique characteristics and evaluation criteria of LLMs, these findings suggest that the fine-tuned model offers greater reliability and accuracy in various real-world applications. This can potentially enhance its applicability across different practical uses. For the detailed results of LLM evaluations, please refer to the Appendix.

\section{Discussion}

In this study, we aimed to enhance the performance of the text-to-image model by fine-tuning the Stable Diffusion 3 model using the DreamBooth technique. While the experimental results were positive, several limitations remain.

Firstly, some abnormal results still appeared during the image generation process. For instance, errors such as an excessive number of fingers or abnormally bent legs were observed in the generated images. These issues indicate the need to improve the model’s detailed representation capabilities.

Additionally, it is noteworthy that the evaluation results from large language models (LLMs) differed significantly from human evaluations. ChatGPT 4.0 and Claude 3.5 rated the images generated by the untuned model as more natural in most cases, whereas actual users preferred the images from the fine-tuned model. This suggests that LLMs and humans may have different visual evaluation criteria, highlighting the need for improved evaluation methods that take this into account in future research.

In terms of quantitative evaluation, we used PSNR, SSIM, and FID metrics to assess model performance. These metrics were useful for measuring image quality and similarity, but it is necessary to introduce newer quantitative evaluation measures to comprehensively assess model performance. For example, metrics that better reflect human cognitive evaluation or more precise evaluation methods for various image attributes could be considered.

\section{Conclusions}

In this study, we aimed to enhance the performance of the text-to-image model by fine-tuning the Stable Diffusion 3 model using the DreamBooth technique. Specifically, experiments were conducted using the "lying on the grass/street" prompt, proposing a method to reduce the generation of abnormal images through fine-tuning.

The experimental results showed that the fine-tuned model significantly reduced the generation of abnormal images in visual evaluations. User surveys also indicated a preference for images generated by the fine-tuned model. Most participants found the images from the fine-tuned model to be more natural, suggesting that the fine-tuned model can provide more natural and consistent images for actual users. Conversely, the evaluations by large language models (LLMs) yielded different results. ChatGPT 4.0 and Claude 3.5 sonet rated the images generated by the untuned model as more natural for most prompts, indicating that LLMs might have different evaluation criteria compared to human visual preferences.

In objective metrics such as Structural Similarity Index (SSIM) and Peak Signal-to-Noise Ratio (PSNR), the fine-tuned model also showed improved results. SSIM measures the similarity between two images, and PSNR indicates the quality of image reconstruction. Both metrics confirmed that the fine-tuned model performed better. However, some abnormal results still appeared during the image generation process, such as excessive numbers of fingers or abnormally bent legs. These issues indicate the need to improve the model’s detailed representation capabilities.

These findings confirm that fine-tuning using DreamBooth is effective in enhancing the performance of text-to-image models. By reducing abnormal image generation, this method can significantly contribute to improving the reliability and accuracy of the models.

This study is expected to contribute to increasing the practicality of text-to-image models and enhancing their applicability in various real-world applications. Future research should explore methods to generate more consistent and high-quality images by applying fine-tuning techniques to various prompts and domains. Additionally, it will be important to address the differences between LLM evaluation criteria and actual user evaluations. Introducing new quantitative evaluation metrics to comprehensively assess model performance could also be considered.

In conclusion, this study serves as a stepping stone towards overcoming the limitations of text-to-image models and developing models with better performance. Through comprehensive evaluation using various metrics and user feedback, we have confirmed the potential for advancement in text-to-image model performance.


%

{\small

\bibliographystyle{ieeetr}

\bibliography{egbib}

}

\appendix
\section*{Appendix}
\section{Training Data and Model Output Examples}

\begin{figure}[h!]
    \centering
    \includegraphics[width=1\linewidth]{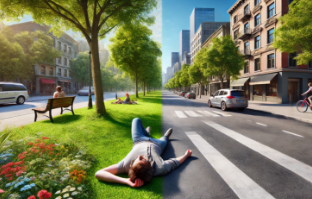}
    \caption{Not Effective Train Data Example}
    \label{fig:enter-label}
\end{figure}

Figure 1 shows an example of training data that was not suitable for learning. Training was not effective when the images was complex, contained multiple people, or when the proportion of the person in the image was small.

\clearpage

\begin{figure}[h!]
    \centering
    \includegraphics[width=0.7\linewidth]{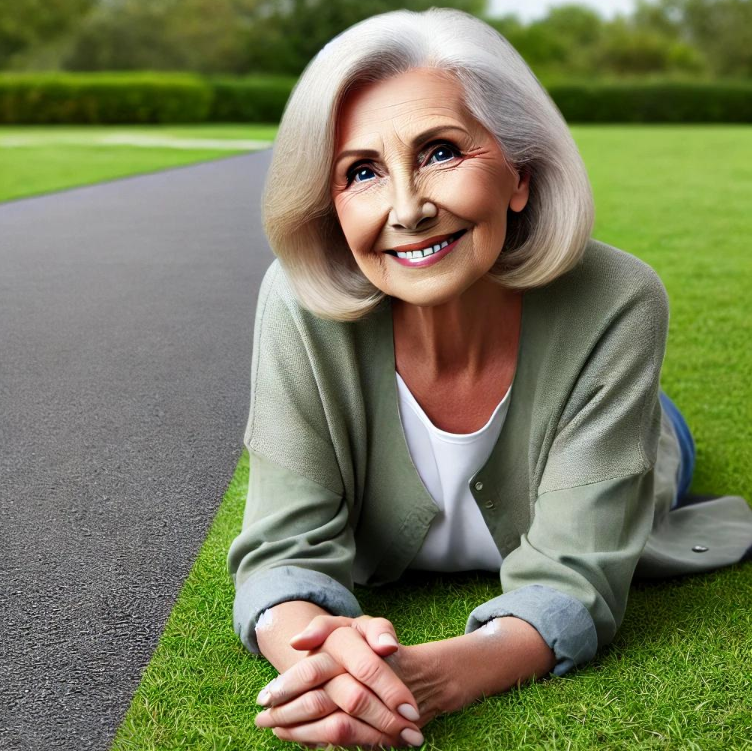}
    \caption{Effective Train Data Example}
    \label{fig:enter-label}
\end{figure}

Figure 2 shows an example of training data that was suitable for learning. The image contains a single person and have a less complex background compared to Figure 1. Additionally, the proportion of the person in the image is much larger than in Figure 1. The images we aimed to correct and obtain through fine-tuning in this study were similar to this example.

\begin{figure}[h!]
    \centering
    \includegraphics[width=0.7\linewidth]{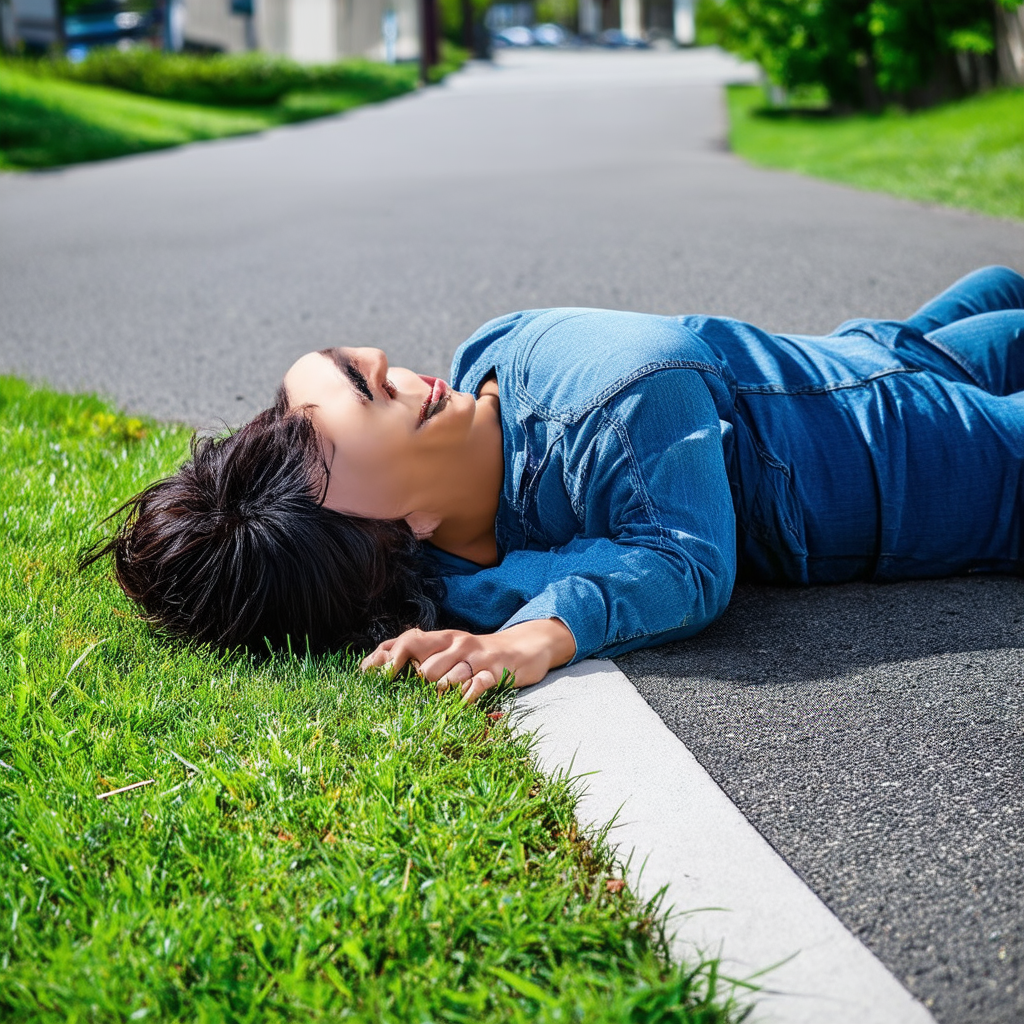}
    \caption{Output Example From Stable Diffusion 3 Model}
    \label{fig:enter-label}
\end{figure}

Figure 3 shows an example of image generated by the Stable Diffusion 3 model without fine-tuning. It can be seen that the human form is generated very abnormally. The shape appears somewhat distorted.

\begin{figure}[h!]
    \centering
    \includegraphics[width=0.7\linewidth]{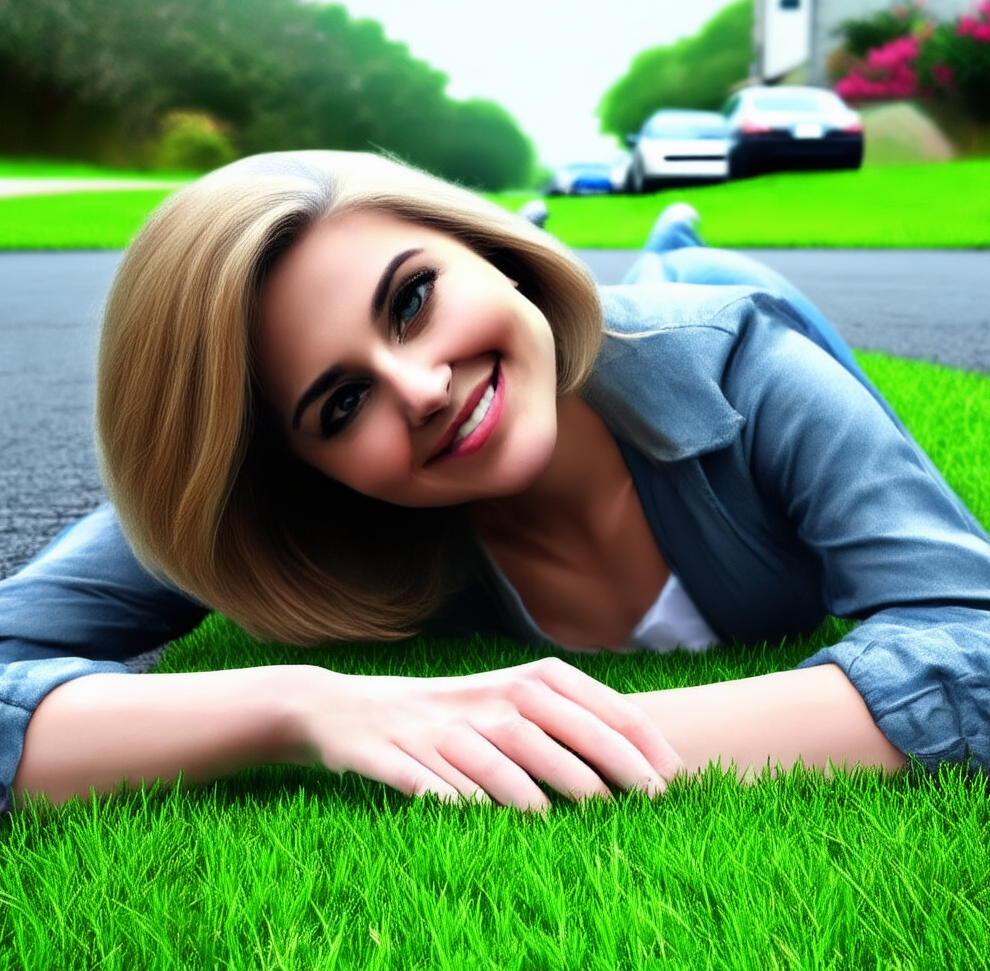}
    \caption{Output Example From Fine-tuned Model}
    \label{fig:enter-label}
\end{figure}

Figure 4 shows an example of image generated by the fine-tuned model. The image resembles the training data and does not appear distorted. However, some abnormal features, such as an excessive number of fingers, still persist. In fact, the training data generated by DALL-E also exhibit slightly abnormal fingers, as shown in Figure 2. It is anticipated that prompting DALL-E to produce more natural shapes for fingers and limbs could result in better training data. Training with such improved data is expected to provide a significant correction effect for abnormal images.

\section{Prompt Examples, and Survey Outcomes}

\begin{figure}[h!]
    \centering
    \includegraphics[width=1\linewidth]{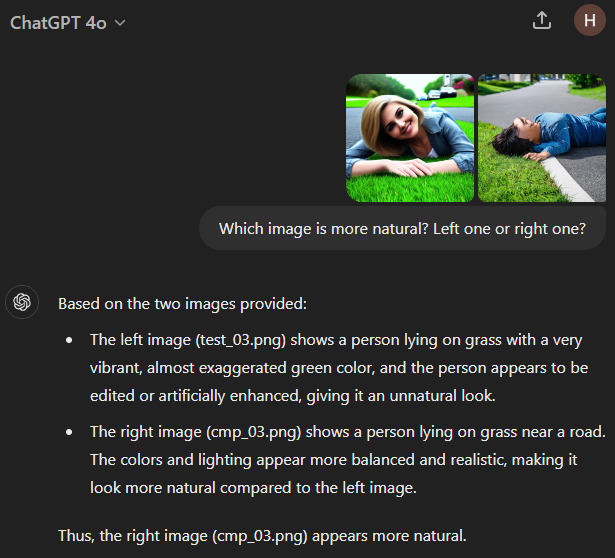}
    \caption{Chat GPT 4o Prompt Example}
    \label{fig:enter-label}
\end{figure}

Figure 5 presents one of the prompts used to evaluate the images generated by the Chat GPT 4.0 model. The results indicated that the images generated by the untuned Stable Diffusion 3 model were judged to be more natural than those generated by the fine-tuned model. This outcome is somewhat intriguing as it contrasts with human evaluations.

\clearpage
\begin{figure}[h!]
    \centering
    \includegraphics[width=1\linewidth]{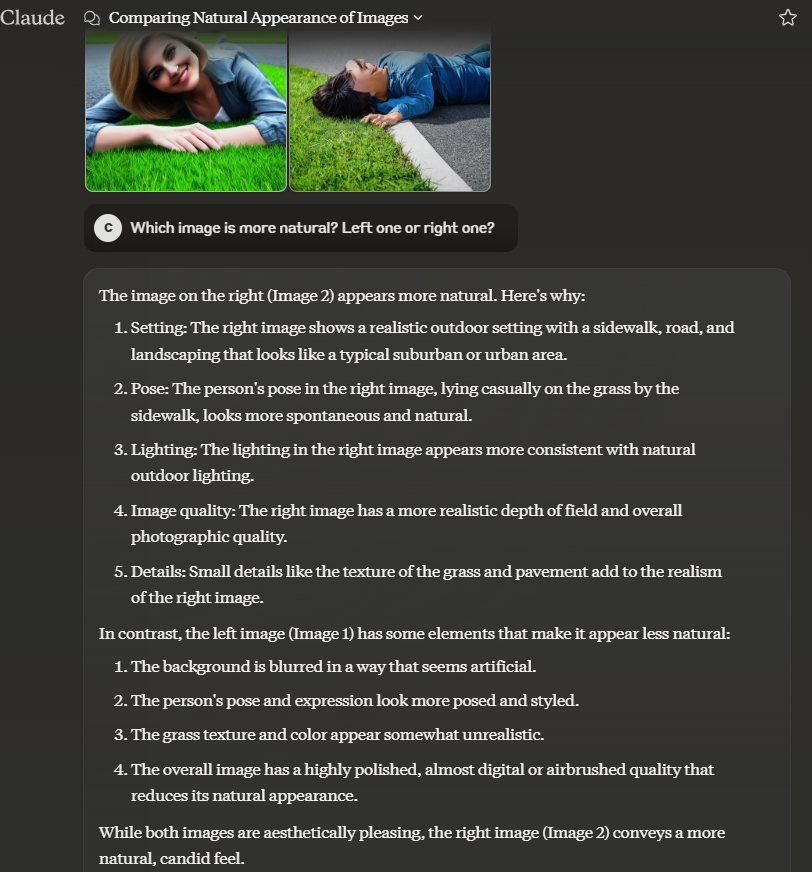}
    \caption{Claude 3.5 Sonet Prompt Example}
    \label{fig:enter-label}
\end{figure}

Figure 6 presents one of the evaluation results from the Claude 3.5 Sonet model. This LLM also evaluated the abnormal images generated by the untuned model as more natural, contrary to human assessments.

\begin{figure}[h!]
    \centering
    \includegraphics[width=1.0\linewidth]{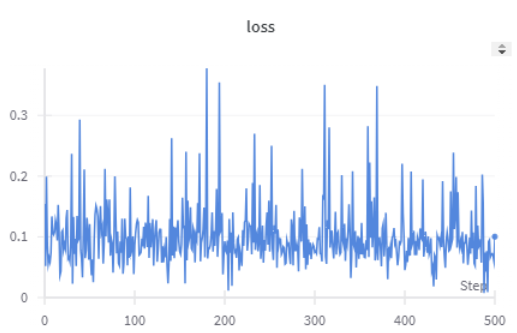}
    \caption{Training Loss Graph}
    \label{fig:enter-label}
\end{figure}

Stable Diffusion, particularly with its complex data and model architecture, is known for having fluctuating loss values during training. During the fine-tuning process of this study, the loss graph also exhibited significant volatility, as shown in Figure 7. However, despite the trend of this graph, the generated images were evaluated and it was determined that the training was relatively successful.

Table 2 presents the survey results from participants. Most people rated the generated results from the fine-tuned model as more natural in at least 8 out of 10 prompts. The generated images evaluated through the survey and LLM are included in the experiments directory on GitHub.

\begin{table*}[htbp]
\centering
\scriptsize

\label{table:survey_results}
\begin{tabular}{|c|c|c|c|c|c|c|c|c|c|c|}
\hline

\textbf{Survey} & \textbf{Participant 1} & \textbf{Participant 2} & \textbf{Participant 3} & \textbf{Participant 4} & \textbf{Participant 5} & \textbf{Participant 6} & \textbf{Participant 7} & \textbf{Participant 8} & \textbf{Participant 9} & \textbf{Participant 10} \\
\hline
1 & left & right & left & left & left & left & left & left & left & left \\
\hline
2 & left & right & right & left & right & right & left & left & left & right \\
\hline
3 & left & left & left & left & left & left & left & left & left & left \\
\hline
4 & left & left & left & left & left & left & left & left & left & left \\
\hline
5 & left & left & left & left & left & left & left & left & left & left \\
\hline
6 & left & left & left & left & left & left & left & left & left & left \\
\hline
7 & left & left & left & left & left & left & left & left & left & left \\
\hline
8 & left & left & left & left & left & left & left & left & left & left \\
\hline
9 & left & left & left & left & left & left & left & left & left & left \\
\hline
10 & left & left & left & left & left & left & left & left & left & left \\
\hline
\end{tabular}
\caption{Survey Results of Fine-tuned Model (Left) vs Original Stable Diffusion 3 (Right)}
\end{table*}

\end{document}